\documentclass{ecai}
\usepackage{times}
\usepackage{graphicx}
\usepackage{latexsym}
\ecaisubmission   % inserts page numbers. Use only for submission of paper.
\usepackage[labelsep=period]{caption}
\usepackage{hyperref}       % hyperlinks
\usepackage{algorithm}
\usepackage{algpseudocode}
\usepackage{amsmath}
\usepackage{amssymb}
\usepackage{enumitem}
\usepackage{multirow}
\usepackage{subcaption}
\usepackage{makecell}

\algnewcommand{\Initialize}[1]{%
  \State \textbf{Initialize:}
  \Statex \hspace*{\algorithmicindent}\parbox[t]{.8\linewidth}{\raggedright #1}
}
\algnewcommand{\Input}[1]{%
  \State \textbf{Input:}
  \Statex \hspace*{\algorithmicindent}\parbox[t]{.8\linewidth}{\raggedright #1}
}
\algnewcommand{\Output}[1]{%
  \State \textbf{Output:}
  \Statex \hspace*{\algorithmicindent}\parbox[t]{.8\linewidth}{\raggedright #1}
}

\DeclareMathOperator*{\logdet}{log\,det}

\DeclareMathOperator*{\Tr}{\,Tr}

\begin{document}

\title{Adversarial Domain Adaptation Being Aware of Class Relationships}

\author{Zeya Wang \institute{Petuum, Inc., United States, email: zw17.rice@gmail.com, \{pengtao.xie, eric.xing\}@petuum.com} \ \institute{Rice University, United States} 
\and
Baoyu Jing \institute{University of Illinois at Urbana-Champaign, United States, email: baoyuj2@illinois.edu} \and
Yang Ni \institute{Texas A\&M University, United States, email: yni@stat.tamu.edu} \and
Nanqing Dong \institute{University of Oxford, United Kingdom, email: nanqing.dong@cs.ox.ac.uk} \and
Pengtao Xie \footnotemark[1]
\and 
Eric Xing \footnotemark[1]
}

\maketitle
\bibliographystyle{ecai}

\begin{abstract}
Adversarial training is a useful approach to promote the learning of transferable representations across the source and target domains, which has been widely applied for domain adaptation (DA) tasks based on deep neural networks. Until very recently, existing adversarial domain adaptation (ADA) methods ignore the useful information from the label space, which is an important factor accountable for the complicated data distributions associated with different semantic classes. Especially, the inter-class semantic relationships have been rarely considered and discussed in the current work of transfer learning. In this paper, we propose a novel relationship-aware adversarial domain adaptation (RADA) algorithm, which first utilizes a single multi-class domain discriminator to enforce the learning of inter-class dependency structure during domain-adversarial training and then aligns this structure with the inter-class dependencies that are characterized from training the label predictor on source domain. Specifically, we impose a regularization term to penalize the structure discrepancy between the inter-class dependencies respectively estimated from domain discriminator and label predictor. Through this alignment, our proposed method makes the adversarial domain adaptation  aware of the class relationships. Empirical studies show that the incorporation of class relationships significantly improves the performance on benchmark datasets.
\end{abstract}

\section{INTRODUCTION}
The success of deep learning largely depends on large-scale datasets with labels (e.g. ImageNet \cite{deng2009imagenet}). Manually annotating labels is costly and time-consuming, which becomes an obstacle for applying deep learning models to new datasets \cite{ganin2016domain}. An effective approach to build a model on unlabeled data in a target domain is to leverage off-the-shelf labeled data from its relevant source domains. However, due to domain shift \cite{torralba2011unbiased}, models trained on source domains usually do not generalize well to target domains. 
  
Recently, adversarial training has been introduced to learn domain-invariant features and substantially improves the domain adaptation performance \cite{ganin2016domain}. These adversarial-learning-based methods incorporate a domain discriminator to encourage domain confusion for minimizing the distribution discrepancy between source and target domains \cite{ganin2016domain,ben2007analysis,long2017deep,tzeng2017adversarial,pei2018multi}. Despite the significant improvement from adversarial domain adaptation, most existing methods simply match the distributions across domains without considering the structure behind the complicated data distributions. The conditional distributions of data given different associated semantic classes can be different, which may lead to multimodal distributions for multi-class classification. Failing to capture the modes of data distribution will mislead the alignment of distributions across domains \cite{pei2018multi,long2018conditional,arora2017generalization}. Current attempts focus on revealing this complex structure within the feature space, but ignore the high-level semantics from the label space. Some method has been proposed to disentangle semantic latent variables from domain latent variables in the latent manifold, but it relies on a variational auto-encoder to reconstruct the latent variables and does not explicitly account for the high-level semantics from the label space  \cite{cai2019learning}. Some recent studies design separate class-wise domain discriminators, where each discriminator is only responsible for the distribution alignment for one semantic class \cite{pei2018multi,chen2017no}. Including the class information, these approaches successfully mitigates the false distribution alignment across domains. However, assigning separate discriminator for each class essentially constrains all classes to be orthogonal with each other. These methods do not explore the inter-class semantic relationships in the label space for DA. 

\begin{figure}[t!]
\centerline{\includegraphics[width=0.9\linewidth]{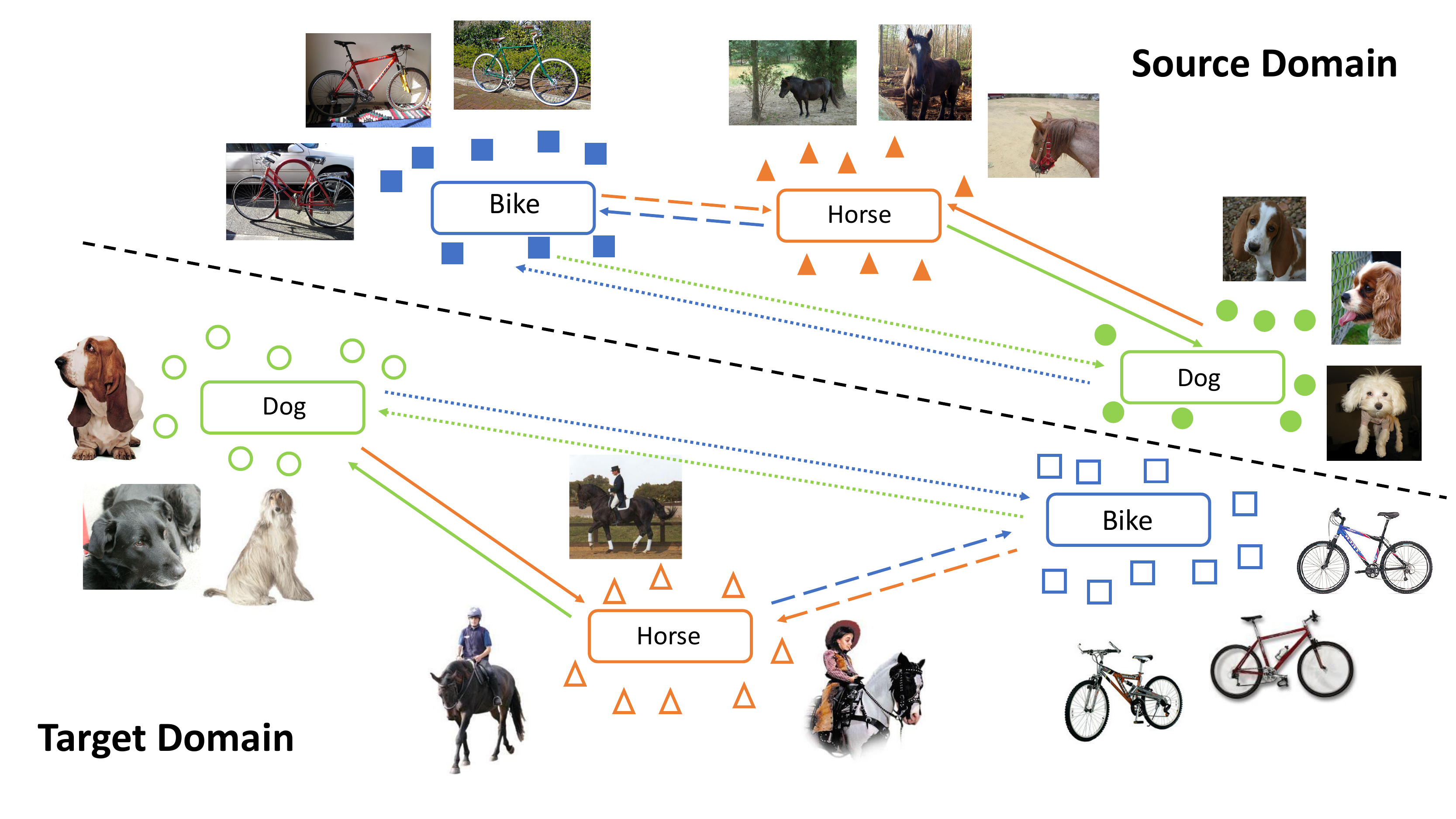}}
   \caption{Class relationships are intuitively supposed to be similar between source and target domains (different line types imply different degree of class relationships). We are motivated to encourage the similarity for making adversarial domain adaptation aware of class relationships.}
   \label{fig:1}
\end{figure}

Utilizing structure information from the label space could be helpful for capturing the multimodal structure more accurately. Intuitively, the class relationships are supposed to remain consistent across domains (Figure \ref{fig:1}), which motivates us to exploit the structure information among semantic classes and inject it into the learning process of DA.  Multiple task learning (MTL) jointly learns multiple related tasks through knowledge sharing, where structure learning has gained growing popularity for explicitly exploiting hidden task structures. Gaussian graphical model is a powerful tool for studying conditional dependency structure among random variables, so has been widely used for learning the structure of task relationships \cite{gonccalves2014multi}. Recently, this approach has been extended to exploit the class relationships with deep neural networks (DNNs) for improved video categorization performance  \cite{jiang2018exploiting}, which provides an effective solution for characterizing inter-class relationships in our work. 

Inspired by this line of work, we first design a single multi-class domain discriminator that implements class-specific domain classification. In doing so, we encourage knowledge sharing across classes for domain classification, which enables the learning of inter-class dependencies, and also favor a parsimonious network. We introduce a structure regularization to constrain the class relationships captured by the domain discrimination maximally agree to the inter-class dependencies that are revealed from label prediction on source domain data. Given that this work focuses on how class relationships are incorporated to improve domain adaptation, we build our model on top of domain adversarial neural network (DANN) \cite{ganin2016domain}, which is the plain adversarial domain adaptation framework. We point out that the presented design and regularizer can be seen as an ``add-on" and be easily integrated to other adversarial domain adaptation frameworks. Experiments on benchmark datasets show the proposed approach outperforms the competing methods. 

%%%%%%%%%%%%%%%Figure 1%%%%%%%%%%%%%%%%%%%%%

\section{RELATED WORK}
\paragraph{Adversarial domain adaptation}
Deep domain adaptation methods attempt to generalize the deep neural networks across different domains. The most commonly used approaches are based on discrepancy minimization \cite{tzeng2014deep,long2015learning,sun2016return,long2017deep,long2016unsupervised,jada2019, jing2018cross} or adversarial training \cite{ganin2015unsupervised,ganin2016domain,tzeng2017adversarial,zhang2018collaborative,dong2018unsupervised}. Adversarial training, inspired by generative modeling in GANs \cite{goodfellow2014generative}, is an important approach for deep transfer learning tasks. DANN \cite{gonccalves2014multi} is proposed with a domain discriminator for classifying whether a sample is from the source or target domains \cite{ganin2016domain,ganin2015unsupervised}. With a gradient reversal layer (GRL), it promotes the learning of discriminative features for classification, and ensures the learned feature distributions over different domains are similar. Recent works realize the importance of exploiting the complex structure behind the data distributions for domain adaptation rather than just aligning the whole source and target distributions \cite{long2018conditional,pei2018multi}. Multi-adversarial domain adaptation(MADA) utilizes the information from the label space by assigning class-wise discriminators to capture the multimodal structure owing to different classes \cite{pei2018multi}. However, the structure information from label space is unexplored for DA.

\paragraph{Structure learning} Multi-task learning (MTL) seeks to improve the generalization performance by transferring knowledge among related tasks. This knowledge sharing feature makes it possible for learning the structure among tasks, so structure learning, which studies how to accurately characterize the task relationships, has become a central issue of MTL \cite{gonccalves2014multi,zhang2010convex}.As one of the earliest MTL models, DNNs also share certain commonalities (neurons of the hidden layer) among the neurons of the output layers \cite{caruana1997multitask,jiang2018exploiting}. Inspired by the methods explicitly modeling task relationships in MTL \cite{gonccalves2016multi,zhang2010convex}, recent studies for multi-class classification using CNNs exploit and harness the inter-class relationships through imposing a regularization, which has been successfully validated for improving the video categorization performance  \cite{jiang2018exploiting}.

\section{METHODS}
In this section, we first discuss how class relationships are modeled with DNNs, followed by the design of single discriminator to perform class-specific domain classification. We then introduce our RADA algorithm that is able to keep the domain adversarial training aware of class relationships.
\subsection{Inter-class dependency structure learning with deep neural networks}
\label{sec:3:1}

For the multi-class classification problem, the data $\mathcal{D} = \{\mathbf{x_m}, \mathbf{y_m}\}_{m=1}^M$ is given, where $\mathbf{x_m}$ represents the input features and $\mathbf{y_m}\in \mathbb{R}^K$ is the associated label for each sample. A DNN $f: \mathbf{x} \mapsto \mathbf{y}$ is used to map the input features of each sample to its associated class $k=1,\cdots, K$ through a large number of interconnected neurons. Typically, these neurons are arranged in multiple layers, e.g., convolutional and pooling layers. In the classification task, a stack of fully connected (FC) layers are often on top of these layers for predicting the final class scores. Only considering the FC layers in a network with $L$ layers in total,  we use $\mathbf{W}^{[l]} \in \mathbb{R}^{N_{l-1} \times N_{l}}$ and $\mathbf{b}^{[l]} \in \mathbb{R}^{N_{l}}$ to  denote the weight matrix and bias vector of neurons in the $l$-th layer respectively, where $N_{l}$ denotes the number of neurons in that layer. Let $\mathbf{a}^{[l-1]}$ and $\mathbf{a}^{[l]}$ denote the input and output of the $l$-th layer with an activation function $g(\cdot)$. We have $\mathbf{a}^{[l]} = g(\mathbf{W}^{[l]^T}\mathbf{a}^{[l-1]} + \mathbf{b}^{[l]})$, and the final output of the network is $\mathbf{\hat y} = f(\mathbf{x}) = \mathbf{a}^{[L]}$. For simplicity of the following discussion, we concatenate $\mathbf{b}^{[l]}$ to the row vectors of $\mathbf{W}^{[l]}$ to have a unified weight matrix $\mathbf{W}^{[l]} \in \mathbb{R}^{(N_{l-1}+1) \times N_{l}}$. The training objective can be calculated through a cross entropy loss $L_y$:
\begin{equation}
    \min \sum_m L_y(f(\mathbf{x_m}), \mathbf{y_m}). 
    \label{eq3:1:1}
\end{equation}

Inspired by recent research for learning task relationships in MTL \cite{zhang2010convex,jiang2018exploiting,gonccalves2016multi}, in classification problems, DNN has been used to exploit the inter-class dependency structure through additional regularization on the output layer to enforce knowledge sharing across different classes.  
One typical way to model the dependency structure among $K$ classes is through a precision matrix $\mathbf{\Omega} \in \mathbb{R}^{K \times K}$, of which each off-diagonal element captures the pairwise partial correlation between classes. Specifically, we assume the row vectors of weight matrix $\mathbf{W}^{[L]}$ of the output layer follow a multivariate Gaussian distribution $\mathbf{W}_i^{[L]} \sim \mathcal{N}(0, \mathbf{\Omega}^{-1}), i = 1, \cdots, N_{L-1}+1$. Let $d = N_{L-1} + 1$. By maximizing the log-likelihood of $\mathbf{\Omega}$ following this assumption of Gaussian distribution  subject to the positive semidefinite constraint (denoted as $\mathbf{\Omega} \succeq 0$), $\mathbf{\Omega}$ can be optimized concurrently with the training objective in equation \eqref{eq3:1:1} by:
\begin{equation}
    \begin{aligned}
     & \min_{\mathbf{\Omega}} - d\logdet(\mathbf{\Omega}) + \Tr(\mathbf{W}^{[L]}\mathbf{\Omega} {\mathbf{W}^{[L]}}^\top), \\
     & s.t. \quad \mathbf{\Omega}  \succeq 0. 
\end{aligned}
    \label{eq3:1:2}
\end{equation}

%%%%%%%%%%%%%%%Figure 2%%%%%%%%%%%%%%%%%%%%%
\begin{figure*}
\centerline{\includegraphics[width=0.8\linewidth]{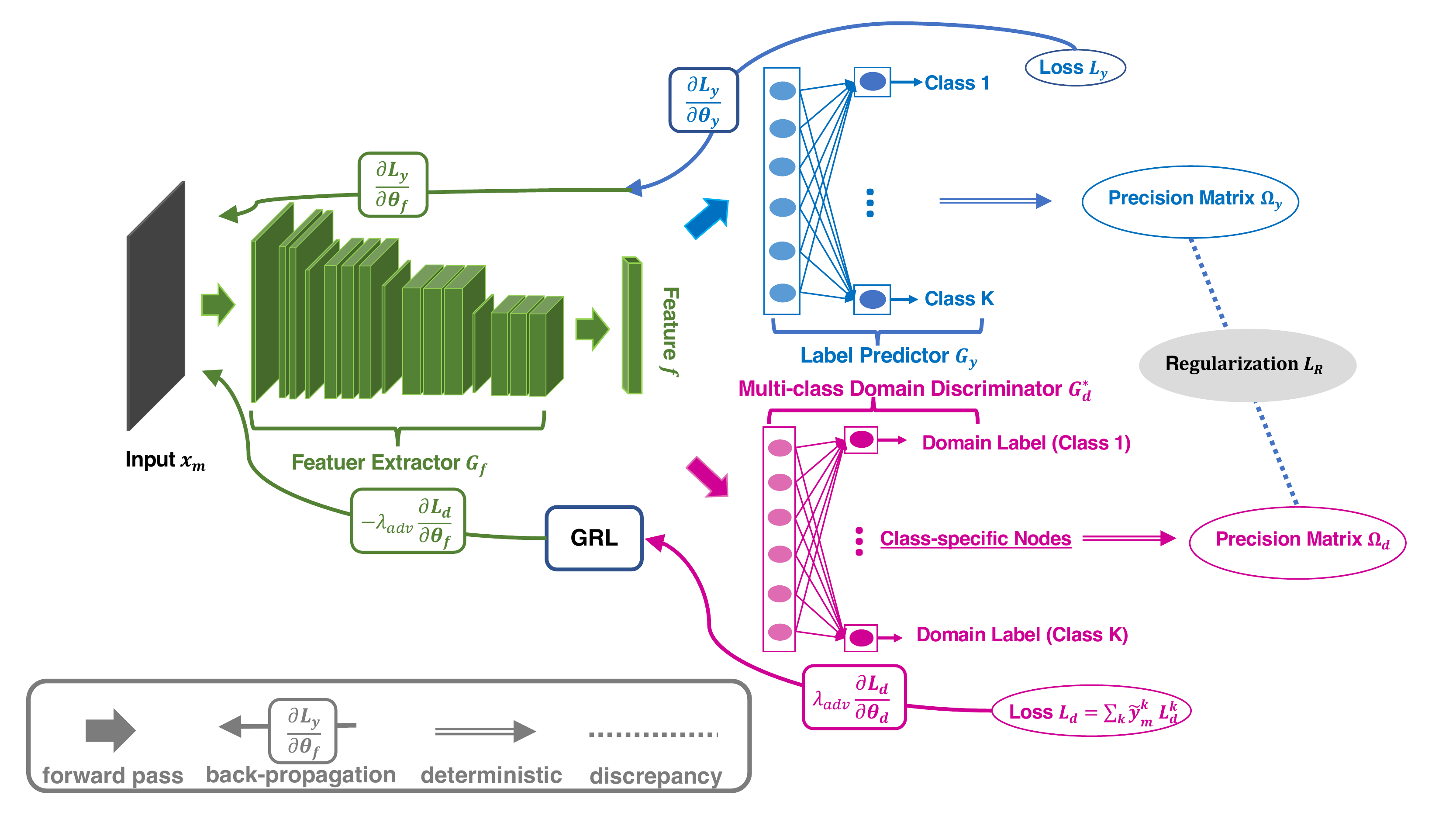}}
   \caption{The architecture of the proposed RADA algorithm built on top of the plain DANN model. In our paper we use a one-layer domain discriminator with $(x \rightarrow 512 \rightarrow K)$. Note that double arrows represent deterministic inference and dashed lines denote the structure discrepancy.}
\label{fig:2}
\end{figure*}
%%%%%%%%%%%%%%%%%%%%%%%%%%%%%%%%%%%%%%%%%%%%%%%%%

\subsection{Multi-class adversarial domain adaptation}
In an \emph{unsupervised domain adaptation}(UDA) problem, we are given labeled source domain data $\mathcal{D}_s = \{\mathbf{x}^s_m, \mathbf{y}^s_m\}_{m=1}^{M_s}$ and unlabeled target domain data $\mathcal{D}_t = \{\mathbf{x}^t_m\}_{m=1}^{M_t}$. DANN has been designed to extract domain invariant features between source and target domains through an adversarial training scheme \cite{ganin2016domain}. The whole architecture consists of three parts: a feature extractor $G_f$, a label predictor $G_y$, and a domain discriminator $G_d$. $G_f$ and $G_y$ together form a standard feed-forward DNN $f(\cdot)$ for predicting class labels. $G_d$ is trained to discriminate samples between source and target domains, while $G_f$ is fine-tuned to confuse $G_d$. Let $\theta_f$, $\theta_y$, and $\theta_d$ denote the parameters of $G_f$, $G_y$, and $G_d$, respectively. In the adversarial training procedure, $\theta_d$ is learned by minimizing a binary cross entropy loss $L_d$ over the domain labels $\mathbf{d}_m$, while $\theta_f$ is learned by maximizing $L_d$ jointly with minimizing $L_y$ (equation \eqref{eq3:2:1}). This is achieved by integrating a gradient reversal layer (GRL) between $G_f$ and $G_d$, finally ensuring the feature distributions over the source and target domains are made similar.

\begin{equation}
\begin{aligned}
&\min \frac{1}{M_s} \sum_{\mathbf{x_m} \in \mathcal{D}_s} L_y(G_y(G_f(\mathbf{x_m})), \mathbf{y_m}) \\
& + \frac{\lambda_{adv}}{M_s + M_t} \sum_{\mathbf{x_m} \in \mathcal{D}_s \cup \mathcal{D}_t} L_d(G_d(\mathcal{R}(G_f(\mathbf{x_m}))), \mathbf{d_m})
\label{eq3:2:1}
\end{aligned}
\end{equation}
where $\mathcal{R(\cdot)}$ is a pseudo-function for GRL \cite{ganin2016domain}, and $\lambda_{adv}$ is a balancing parameter for adversarial loss.

In order to capture the multimodal structure of data distribution that is accountable by different semantic classes for domain adaptation, a design of multiple discriminators has been applied, such that one discriminator is responsible for matching the source and target domain data associated with one certain class \cite{pei2018multi}. This design has been proved to successfully enhance positive transfer and alleviate negative transfer, a phenomenon where the source domain data play a part in the reduced learning performance in the target domain \cite{li2019towards,guo2018general,wei2018learning}. However, there are still two concerns: 1) it has a strong assumption of orthogonality across classes during distribution alignment, i.e., it neglects the structure information among the semantic classes 2) the number of discriminators increased with the number of classes elevates the memory cost for network parameters. In addressing these concerns, we first present a multi-class adversarial domain adaptation  $G^*_d$, where it should be noticed that the way we use ``multi-class'' differs from that in standard multi-class classification. Instead of adopting separate discriminators, we use one single discriminator with a multi-branch design to match the multimodal structure across different classes.

Figure \ref{fig:2} gives a demonstration of $G^*_d$ in the whole network. One shared hidden layer in $G^*_d$ encodes the common discriminative features that can be used to classify domains for all classes. This shared layer is followed by a layer with class-specific nodes, where each node/branch is only responsible for predicting the domain label of the samples from its associated class, and the node will be muted when the samples to be estimated with domain labels are associated with other classes. For one sample, the ground truth of domain label for each class-specific node is consistent with the ground truth of domain label for that sample. We use $L_d^k$ ($k = 1, \cdots, K$) to denote the binary domain classification loss associated with class $k$. With label information $\mathbf{y_m}$, source domain data can be easily assigned to each class-specific node. For the unlabeled target domain data, a weighted sum of loss values from different nodes are calculated, where the probability score vector $\mathbf{\hat y_m}$ given by $G_y$ are used as the weights. Integrating this new design, we update the objective of our multi-class adversarial domain adaptation as:
\begin{equation}
\begin{aligned}
& \min  \frac{1}{M_s} \sum_{\mathbf{x_m} \in \mathcal{D}_s} L_y(G_y(G_f(\mathbf{x_m})), \mathbf{y_m}) \\
& + \frac{\lambda_{adv}}{M_s+M_t} \sum_{k=1}^K \sum_{\mathbf{x_m} \in \mathcal{D}_s \cup \mathcal{D}_t}  \tilde y_m^k L_d^k(G_d^*(\mathcal{R}(G_f(\mathbf{x_m}))), \mathbf{d_m}) \\
\label{eq3:2:2}
\end{aligned}
\end{equation}
where $\mathbf{\tilde y_m} = \{\tilde y_m^k\}_{k=1}^K$ is one-hot encoding of $\mathbf{y_m}$ for $\mathbf{x_m} \in \mathcal{D}_s$ and $\mathbf{\tilde y_m} = \mathbf{\hat y_m}$ for $\mathbf{x_m} \in \mathcal{D}_t$.

\subsection{Adversarial domain adaptation being aware of class relationships}
Incorporating the information of class relationships to the alignment process between the source and target data distributions will relax the orthogonality assumption and help maximally match the multimodal structure of data distributions. Recall that $\mathbf{\Omega}$ is used to model the inter-class dependency structure with DNN from Section \ref{sec:3:1}. By implicitly injecting $\mathbf{\Omega}$ into the adversarial training process, we  make adversarial domain adaptation automatically aware of class relationships. 

With the extracted features from $G_f$, $G_y$ predicts class labels, where the class relationships can be characterized from the prediction at the same time. During the process of domain adaptation, $G^*_d$ aligns the source and target data distributions, which should have similar class structure information. In order for $G^*_d$ to capture a similar inter-class dependency structure during the alignment, the precision matrix $\mathbf{\Omega_d}$, which can be estimated from the class-specific domain classification job implemented by $G^*_d$, is supposed to be consistent with $\mathbf{\Omega_y}$ from the prediction task done by $G_y$. To maximize this consistency, we propose an approach that minimizes the discrepancy between the class relationships respectively learned from $G_y$ and $G^*_d$ (as shown in Figure \ref{fig:2}). Let $\mathbf{\Omega_y}$ and $\mathbf{\Omega_d}$ denote the precision matrices w.r.t the weight matrices $\mathbf{W_y}^{[L]}$ and $\mathbf{W_d}^{[L]}$ of the output layers in $G_y$ and $G^*_d$, thus $\mathbf{\Omega_y}$ and $\mathbf{\Omega_d}$ are used to characterize the inter-class dependencies w.r.t $G_y$ and $G_d$. Let $d_y$ be the value of $d$ calculated w.r.t $G_y$, according to equation \eqref{eq3:1:2}, we can solve $\mathbf{\Omega_y}$ from
\begin{equation}
\begin{aligned}
& \min_{\mathbf{\Omega_y}} - d_y\logdet(\mathbf{\Omega_y}) + \Tr(\mathbf{W_y}^{[L]}\mathbf{\Omega_y} {\mathbf{W_y}^{[L]}}^T);\\
& s.t. \quad \mathbf{\Omega_y}  \succeq 0. 
\end{aligned}
\label{eq3:3:1}
\end{equation}
It is straightforward to derive the solution to this minimization problem, assuming ${\mathbf{W_y}^{[L]}}^T \mathbf{W_y}^{[L]}$ is positive definite. Using the spectral theorem, we can conclude $\mathbf{\Omega_y} = d_y{({\mathbf{W_y}^{[L]}}^T \mathbf{W_y}^{[L]}})^{-1}$ by assuming the positive definiteness of $\mathbf{{W_y}^{[L]}}^T\mathbf{W_y}^{[L]}$(the derivation is included in the appendix). In practice, this positive definiteness is usually guaranteed given the much larger number of hidden nodes compared with the number of classes. In a rare case when it cannot be satisfied, a shrinkage approach can be applied, e.g., adding a small multiple of the identity matrix. Similarly, assuming the positive definiteness of $\mathbf{{W_d}^{[L]}}^T\mathbf{W_d}^{[L]}$ solving
  \begin{equation}
  \begin{aligned}
  & \min_{\mathbf{\Omega_d}}- d_d\logdet(\mathbf{\Omega_d}) + \Tr(\mathbf{W_d}^{[L]}\mathbf{\Omega_d} {\mathbf{W_d}^{[L]}}^T); \\
  & s.t. \quad \mathbf{\Omega_d}  \succeq 0.
  \end{aligned}
  \label{eq3:3:2}
  \end{equation}
  With $d_d$ being the value of $d$ from $G^*_d$, we obtain $\mathbf{\Omega_d} = d_d{({\mathbf{W_d}^{[L]}}^T \mathbf{W_d}^{[L]}})^{-1}$.
    We adopt a structure regularization approach that minimizes the discrepancy between precision matrices $\mathbf{\Omega_y}$ and $\mathbf{\Omega_d}$ (a.k.a. KL divergence \cite{cui2016covariance}). It is an asymmetric metric, thus we can formulate the discrepancy in two ways, as:
      \begin{equation}
    D_{KL}(\mathbf{\Omega_y} || \mathbf{\Omega_d}) = \Tr(\mathbf{\Omega_y}^{-1} \mathbf{\Omega_d}) - \logdet(\mathbf{\Omega_y}^{-1} \mathbf{\Omega_d}) - K
    \label{eq3:3:3}
    \end{equation}
    , which minimizes the divergence from $\mathbf{\Omega_d}$ to $\mathbf{\Omega_y}$, or 
    \begin{equation}
    D_{KL}(\mathbf{\Omega_d} || \mathbf{\Omega_y}) = \Tr(\mathbf{\Omega_d}^{-1} \mathbf{\Omega_y}) - \logdet(\mathbf{\Omega_d}^{-1} \mathbf{\Omega_y}) - K.
    \label{eq3:3:4}
    \end{equation}
    , which minimizes the divergence from $\mathbf{\Omega_y}$ to $\mathbf{\Omega_d}$.
    Inserting $\mathbf{\Omega_y} = d_y{({\mathbf{W_y}^{[L]}}^T \mathbf{W_y}^{[L]}})^{-1}$ and $\mathbf{\Omega_d} = d_d{({\mathbf{W_d}^{[L]}}^T \mathbf{W_d}^{[L]}})^{-1}$ into equation \eqref{eq3:3:3} or \eqref{eq3:3:4}, we design a regularization to minimize the discrepancy of class relationships between $G_y$ and $G^*_d$ in two ways:
\begin{equation}
      \begin{split}
      & \mathbf{d \rightarrow y}: L_R = \Tr({\mathbf{W_y}^{[L]}({\mathbf{W_d}^{[L]}}^T \mathbf{W_d}^{[L]}})^{-1}{\mathbf{W_y}^{[L]}}^T)\\
      - & \frac{d_y}{d_d}\{\logdet({\mathbf{W_y}^{[L]}}^T \mathbf{W_y}^{[L]}) - \logdet({\mathbf{W_d}^{[L]}}^T \mathbf{W_d}^{[L]})\}
      \end{split}
      \label{eq3:3:5}
      \end{equation}
      \begin{equation}
      \begin{split}
      & \mathbf{y \rightarrow d}:  L_R = \Tr({\mathbf{W_d}^{[L]}({\mathbf{W_y}^{[L]}}^T \mathbf{W_y}^{[L]}})^{-1}{\mathbf{W_d}^{[L]}}^T) \\
      - & \frac{d_d}{d_y} \{\logdet({\mathbf{W_d}^{[L]}}^T \mathbf{W_d}^{[L]}) - \logdet({\mathbf{W_y}^{[L]}}^T \mathbf{W_y}^{[L]})\}
      \end{split}
      \label{eq3:3:6}
      \end{equation}
      Given the asymmetry of the KL divergence between matrices, we will implement both of $D_{KL}(\mathbf{\Omega_d} || \mathbf{\Omega_y})$ and $D_{KL}(\mathbf{\Omega_y} || \mathbf{\Omega_d})$ in our experiments and investigate their difference in our following discussion. Integrating the penalty $L_R$ to equation \eqref{eq3:2:2}, we have our final training objective:
        \begin{equation}
      \begin{split}
      \min & \frac{1}{M_s} \sum_{\mathbf{x_m} \in \mathcal{D}_s} L_y(G_y(G_f(\mathbf{x_m})), \mathbf{y_m}) + \lambda_R L_R \\
      + & \frac{\lambda_{adv}}{M_s+M_t} \sum_{k=1}^K \sum_{\mathbf{x_m} \in \mathcal{D}_s \cup \mathcal{D}_t}  \tilde y_m^k L_D^k(G_d^*(\mathcal{R}(G_f(\mathbf{x_m}))), \mathbf{d_m}) 
      \end{split}
      \label{eq3:3:7}
      \end{equation}
      where $\lambda_R$ is a balancing parameter for the relationship-aware regularization term. 

\section{EXPERIMENTS}
%%%%%%%%%%%%%%%%%%%Table 1%%%%%%%%%%%%%%%%%%%
\begin{table*}[t!]
\begin{center}
{\caption{Mean accuracy (\%) on \textit{ImageCLEF-DA} for UDA (\textit{ResNet-50})}
\label{tab:1}}
\begin{tabular}{cccccccc}
\hline
Method & I$\rightarrow$P & P$\rightarrow$I & I$\rightarrow$C & C$\rightarrow$I & C$\rightarrow$P & P$\rightarrow$C & Average\\
\hline 
ResNet \cite{he2016deep} &74.8  &83.9 & 91.5 & 78.0 & 65.5 &  91.2 & 80.7\\ 
DAN \cite{long2015learning} & 75.0 &86.2 &93.3  &84.1  & 69.8 & 91.3 &83.3 \\ 
RTN \cite{long2016unsupervised} &75.6  &86.8 & 95.3 &86.9  & 72.7 &92.2  & 84.9\\ 
DANN \cite{ganin2016domain} &75.0  &86.0 &96.2  & 87.0 & 74.3 &91.5  &85.0 \\ 
JAN \cite{long2017deep} & 76.8 &88.0 & 94.7 &89.5  &74.2  &91.7  &85.8 \\ 
CAN \cite{zhang2018collaborative} & 78.2 &87.5 & 94.2 &89.5  &75.8  &89.2  &85.7 \\ 
MADA \cite{pei2018multi} &75.0  &87.9 & 96.0 &88.8  &75.2  &92.2  &85.8 \\ 
\hline 
\textbf{RADA$_{y\rightarrow d}$} &  78.8& 92.1 & 97.3 & 90.9 &76.4  &94.6  & 88.4\\ 
\textbf{RADA$_{d\rightarrow y}$} & \textbf{79.2} & \textbf{92.4}& \textbf{97.5} &\textbf{91.1}  & \textbf{76.6} & \textbf{95.3} & \textbf{88.7}\\ 
\hline
\end{tabular}
\end{center}
\end{table*}
%%%%%%%%%%%%%%%%%%%Table 2%%%%%%%%%%%%%%%%%%%
  \begin{table*}[t!]
\begin{center}
{\caption{Mean accuracy (\%) on \textit{Office-31} for UDA (\textit{ResNet-50})}
\label{tab:2}}
\begin{tabular}{cccccccc}
\hline
Method & A$\rightarrow$W & D$\rightarrow$W & W$\rightarrow$D & A$\rightarrow$D & D$\rightarrow$A & W$\rightarrow$A & Average\\
\hline 
ResNet \cite{he2016deep} & 68.4 & 96.7&99.3  &68.9  &62.5  &60.7  &76.1 \\ 
TCA \cite{pan2011domain} &74.7  & 96.7& 99.6 & 76.1 & 63.7 &62.9  & 79.3\\ 
GFK \cite{gong2012geodesic} & 74.8 &95.0 & 98.2 &76.5  &65.4  & 63.0 & 78.8\\
DDC \cite{tzeng2014deep} & 75.8 &95.0 &98.2  &77.5  &67.4  &64.0  &79.7 \\ 
DAN \cite{long2015learning} &83.8  & 96.8& 99.5 &78.4  &66.7  &62.7  &81.3 \\ 
RTN \cite{long2016unsupervised} & 84.5 &96.8 & 99.4 &77.5  &66.2  &64.8  &81.6 \\ 
DANN \cite{ganin2016domain} &82.0  &96.9 &99.1  &79.7  &68.2  & 67.4 &82.2 \\ 
ADDA \cite{tzeng2017adversarial} & 86.2 &96.2 &98.4  &77.8  &69.5  &68.9  &82.9 \\ 
JAN \cite{long2017deep} & 85.4 &97.4 & 99.8 &84.7  &68.6  &70.0  &84.3 \\ 
JDDA \cite{jada2019} & 82.6 &95.2 & 99.7 &79.8  &57.4  &66.7  &80.2 \\ 
CAN \cite{zhang2018collaborative} & 81.5 &98.2 & 99.7 &85.5  &65.9  &63.4  &82.4 \\ 
MADA \cite{pei2018multi} &90.0  &97.4 & 99.6 &87.8  & 70.3 &66.4  &85.2 \\ 
\hline 
\textbf{RADA$_{y\rightarrow d}$} &\textbf{91.5}  & \textbf{99.0} & \textbf{100.0} & 90.3 & \textbf{71.5} & 70.1 & 87.1\\ 
\textbf{RADA$_{d\rightarrow y}$} &\textbf{91.5}  &98.9 & \textbf{100.0} & \textbf{90.7} & \textbf{71.5} & \textbf{71.3} & \textbf{87.3}\\ 
\hline
\end{tabular}
\end{center}
\end{table*}

%%%%%%%%%%%%%%%%%%%%%%%%%%%%%%%%%%%%%
\subsection{Experiment setup}
\paragraph{Datasets}
We evaluate our model performance on two benchmarks.
The first dataset is \textit{ImageCLEF-DA}{~\footnote{http://imageclef.org/2014/adaptation}}. 
All the images are collected from three public datasets: \textit{Caltech256} ({\bf C}), \textit{ImageNet ILSVRC 2012} ({\bf I}), and \textit{Pascal VOC 2012} ({\bf P}). 
They are in 12 common categories shared by the three datasets, with $50$ images in each category. We evaluate our method on the transfer tasks with all domain combinations: $\mathbf{I}\rightarrow\mathbf{P}$, $\mathbf{P}\rightarrow\mathbf{I}$, $\mathbf{I}\rightarrow\mathbf{C}$,  $\mathbf{C}\rightarrow\mathbf{I}$, $\mathbf{C}\rightarrow\mathbf{P}$ and $\mathbf{P}\rightarrow\mathbf{C}$.  The other dataset is \textit{Office-31} \cite{saenko2010adapting}, which consists of totally $4,110$ images from 31 categories. All the images are collected from three different domains: \textit{Amazon} ({\bf A}), which are downloaded from amazon.com, \textit{DSLR} ({\bf D}), which are taken by digital SLR camera and \textit{Webcam} ({\bf W}), which are recorded with a simple webcam. This dataset with images from different photographical settings represent visual domain shifts. We evaluate our method in terms of classification accuracy on all the six transfer tasks  $\mathbf{A}\rightarrow\mathbf{W}$, $\mathbf{D}\rightarrow\mathbf{W}$, $\mathbf{W}\rightarrow\mathbf{D}$,  $\mathbf{W}\rightarrow\mathbf{D}$, $\mathbf{A}\rightarrow\mathbf{D}$ and $\mathbf{D}\rightarrow\mathbf{A}$. 

%%%%%%%%%%%%%%%%%%%Table 3%%%%%%%%%%%%%%%%%%%
 \begin{table*}[t!]
\begin{center}
{\caption{Mean accuracy (\%) on \textit{Office-31} for PDA from 31 classes to 10 classes (\textit{ResNet-50})}
\label{tab:3}}
\begin{tabular}{cccccccc}
\hline
Method & A$\rightarrow$W & D$\rightarrow$W & W$\rightarrow$D & A$\rightarrow$D & D$\rightarrow$A & W$\rightarrow$A & Average\\
\hline 
ResNet \cite{he2016deep} & 54.5 & 94.6&94.3  &65.6  &73.2  &71.7  &75.6 \\ 
DAN \cite{long2015learning} &46.4  & 53.6& 58.6 & 42.7 & 65.7 &65.3  & 55.4\\ 
ADDA \cite{tzeng2017adversarial} & 43.7 &46.5 &40.1  &43.7  &42.8  &46.0  &43.8 \\ 
RTN \cite{long2016unsupervised} &75.3 &  97.1& 98.3 &66.9  &85.6  &85.7  &84.8 \\ 
JAN \cite{long2017deep} & 43.4 &53.6 & 41.4 &35.7  &51.0  &51.6  &46.1 \\ 
DANN \cite{ganin2016domain} & 41.4 &46.8 & 38.9 &41.4  &41.3  & 44.7 & 42.4\\ 
MADA \cite{pei2018multi} \footnotemark{} & 63.5 & 85.1&  \textbf{99.7}& 67.7 & 59.1 &  63.9& 73.2 \\ 
\hline 
\textbf{RADA$_{y\rightarrow d}$} &\textbf{83.0}  & \textbf{97.4} &97.2 &
  \textbf{87.4} & 86.0 & 85.5 & 89.4\\
\textbf{RADA$_{d\rightarrow y}$} &82.8  & \textbf{97.4} & 97.6 &86.8  & \textbf{86.6} & \textbf{86.3} & \textbf{89.6}\\ 
\hline
\end{tabular}
\end{center}
\end{table*}\footnotetext{Reimplementation.}
%%%%%
\paragraph{Impementation details} The training and testing are implemented by \textbf{Pytorch}.  Among all the transfer tasks, we use stochastic gradient descent (SGD) with momentum of $0.9$ \cite{sutskever2013importance} for minimizing the loss function given by equation \eqref{eq3:3:7}. In \textit{Office-31}, we adopt balanced sampling between classes to increase the chance that samples from each category can be drawn in each batch. The learning rate is initialized with $0.0005$ for all the CNN layers and $0.005$ for all the fully connected layers, and then exponentially decayed during SGD by a factor $(1 + \alpha p)^\beta$, where $\alpha=10$, $\beta=0.75$ and $p \in [0,1]$ is the training progress measured by epoch numbers \cite{ganin2016domain}. All weights and biases are regularized by a weight decay with $L_2$ penalty multiplier set to $0.0005$. $\lambda_{adv}$ for adversarial training is fixed with $1$, decayed with a factor $\frac{1-exp(-10p)}{1+exp(-10p)}$ through the training process, while $\lambda_R$ is fixed with $0.01$ across all the experiments \cite{pei2018multi}. We implement our method based on the \textit{ResNet-50} \cite{he2016deep} pre-trained on the ImageNet dataset as is done in the compared deep learning methods \cite{deng2009imagenet,long2018conditional, pei2018multi}.

\paragraph{Baselines}
We follow standard evaluation protocols for UDA using all labeled source samples and all unlabeled target samples, and report the mean classification accuracy over three random experiments \cite{long2018conditional, pei2018multi}. We compare RADA with recent state-of-the-art deep transfer learning methods based on \textit{ResNet-50}: Deep Domain Confusion (DDC) \cite{tzeng2014deep}, Deep Adaptation Network (DAN) \cite{long2015learning}, Residual Transfer Network (RTN) \cite{long2016unsupervised}, DANN \cite{ganin2016domain}, Adversarial Discriminative Domain Adaptation (ADDA) \cite{tzeng2017adversarial}, Joint Adaptation Network (JAN) \cite{long2017deep}, MADA \cite{pei2018multi}, Collaborative and Adversarial Network (CAN) \cite{zhang2018collaborative}, and Joint Discriminative Domain Adaptation (JDDA) \cite{jada2019}; and traditional machine learning methods: Transfer Component Analysis (TCA) \cite{pan2011domain}, Geodesic Flow Kernel (GFK) \cite{gong2012geodesic}. 

\begin{figure*}
\centering
\begin{subfigure}[t]{0.24\textwidth}
\centerline{\includegraphics[width=0.9\linewidth]{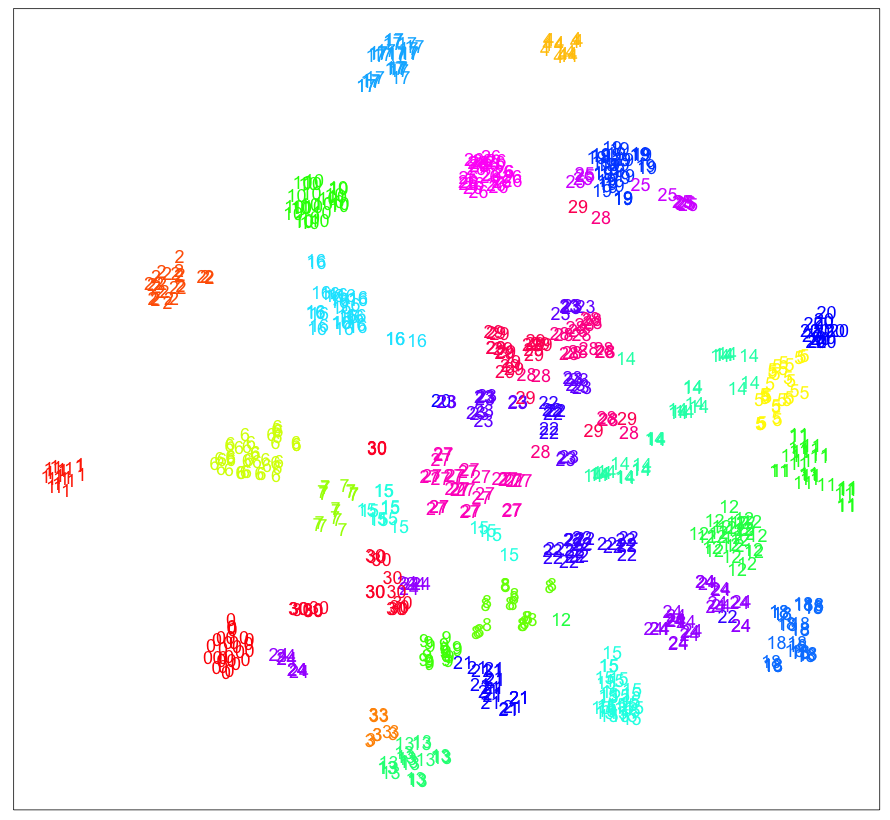}}
\caption{DANN}
\end{subfigure}% 
\begin{subfigure}[t]{0.24\textwidth}
\centerline{\includegraphics[width=0.9\linewidth]{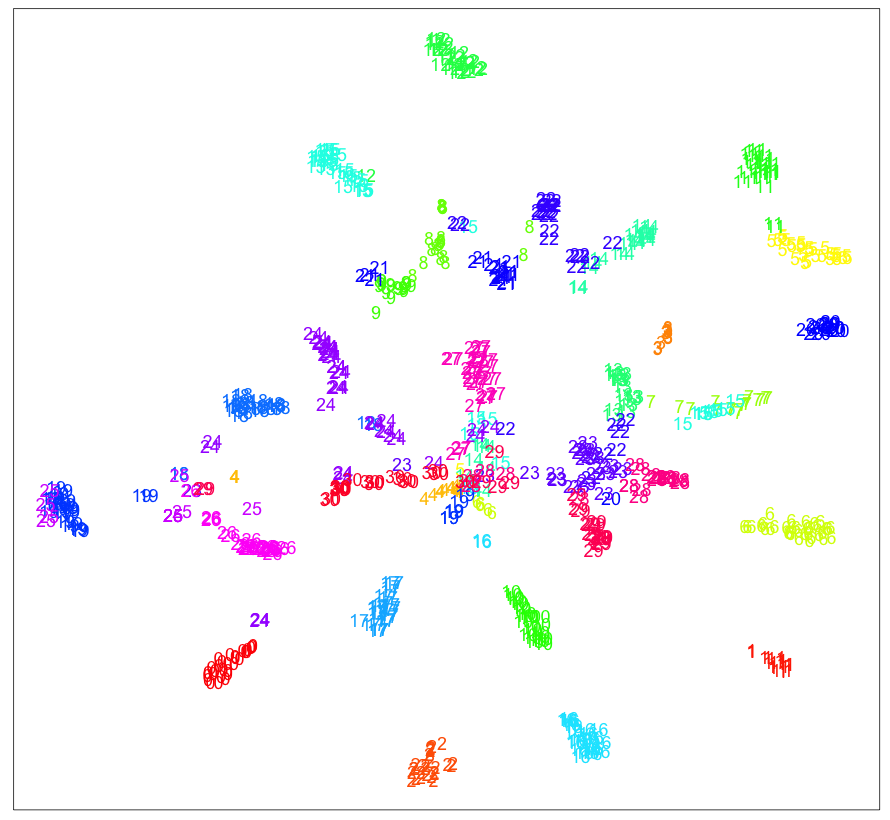}}
\caption{MADA}
\end{subfigure}% 
\begin{subfigure}[t]{0.24\textwidth}
\centerline{\includegraphics[width=0.9\linewidth]{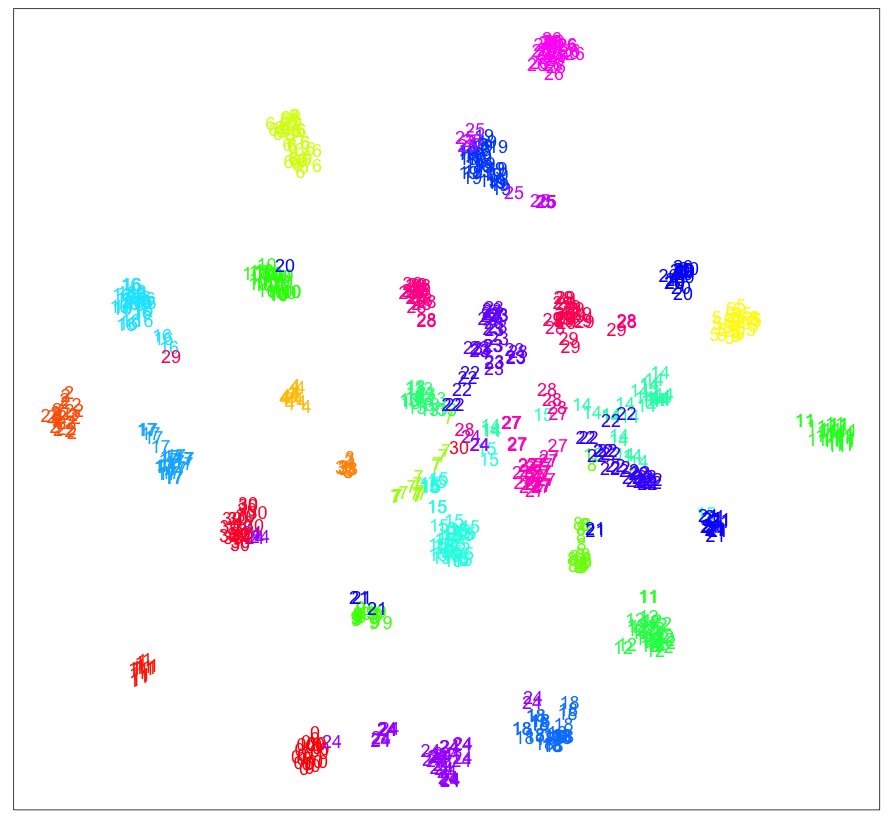}}
\caption{RADA$_{y \rightarrow d}$}
\end{subfigure}% 
\begin{subfigure}[t]{0.24\textwidth}
\centerline{\includegraphics[width=0.9\linewidth]{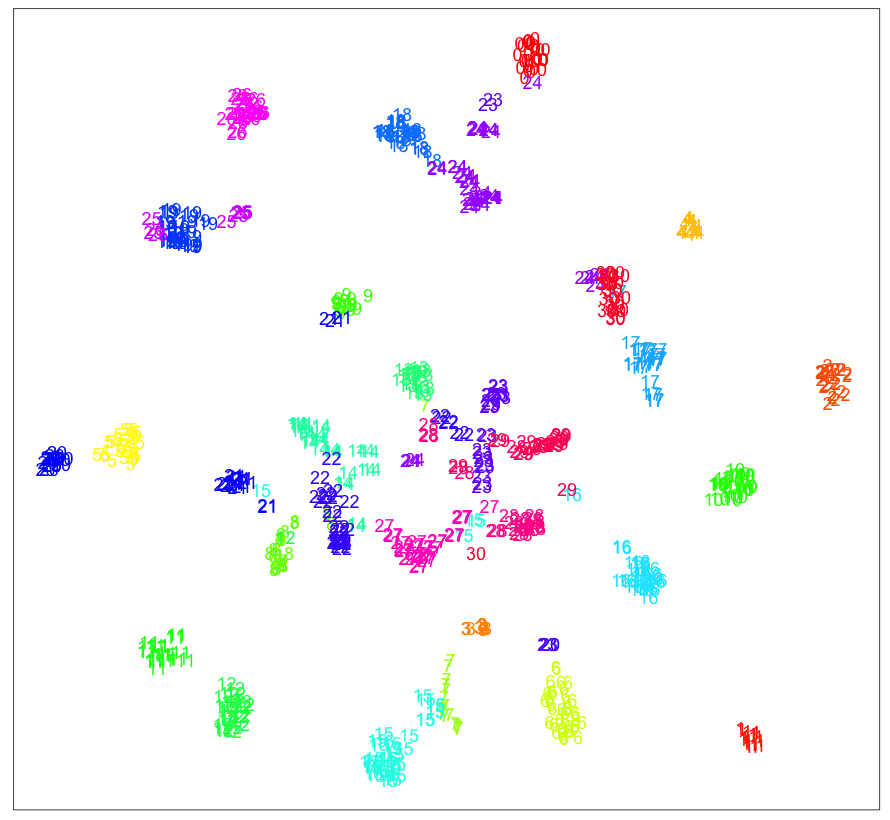}}
\caption{RADA$_{d \rightarrow y}$}
\end{subfigure}% 
\caption{The t-SNE visulization of embedded features from target domain. }
\label{fig:3:1}
\end{figure*}

\begin{figure*}[t]
\centering
\begin{subfigure}[t]{0.35\textwidth}
\centerline{\includegraphics[width=0.8\linewidth]{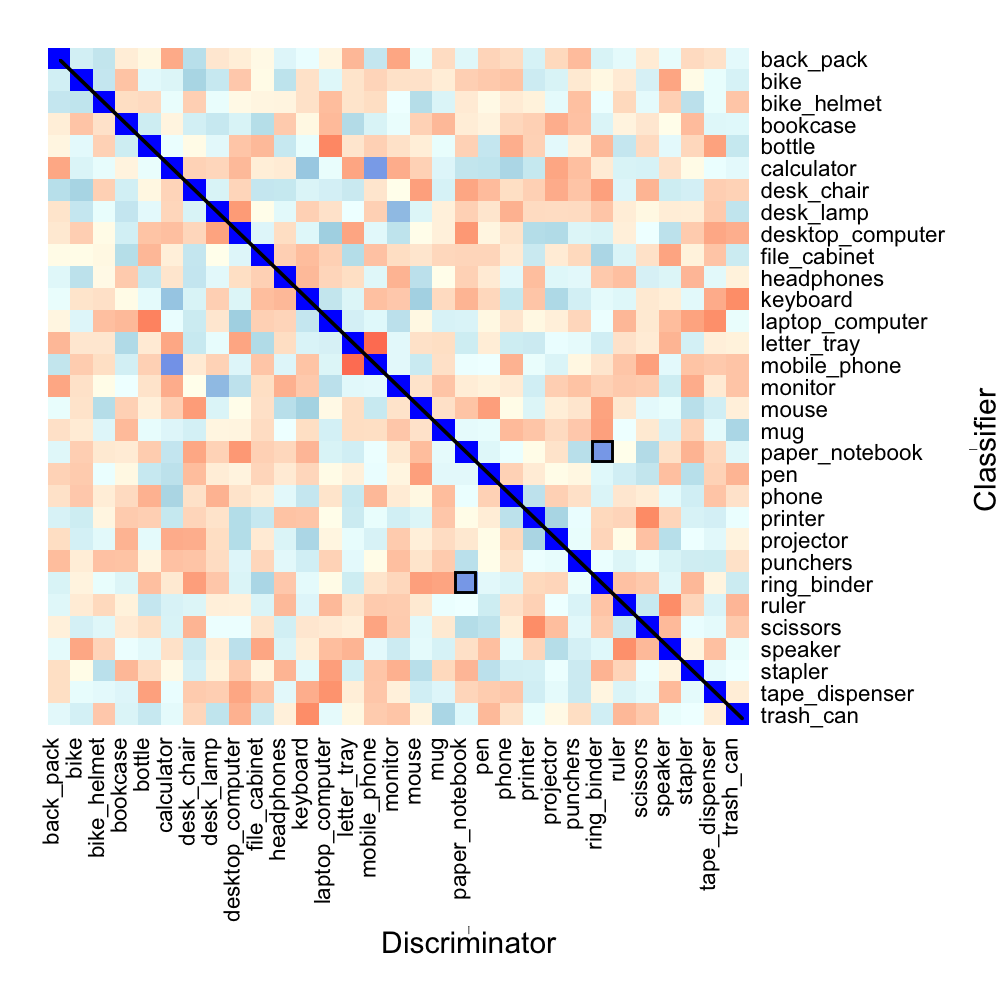}}
\caption{Heatmap of $\hat \rho_{ij}$ (blue: $\geq0$, red: $<0$)}
\label{fig:3:2:1}
\end{subfigure}% 
\begin{subfigure}[t]{0.32\textwidth}
\centerline{\includegraphics[width=0.8\linewidth]{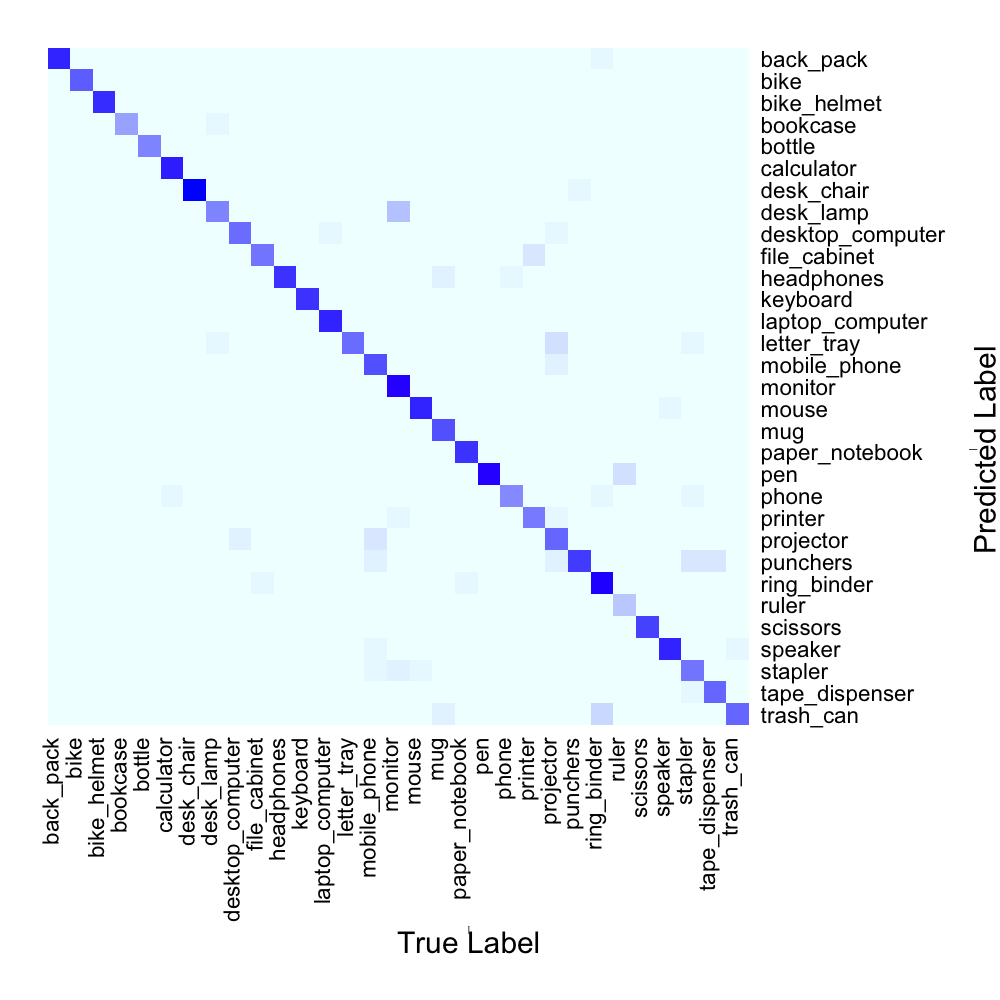}}
\caption{Confusion matrix of RADA$_{d\rightarrow y}$}
\label{fig:3:2:2}
\end{subfigure}% 
\begin{subfigure}[t]{0.32\textwidth}
\centerline{\includegraphics[width=0.8\linewidth]{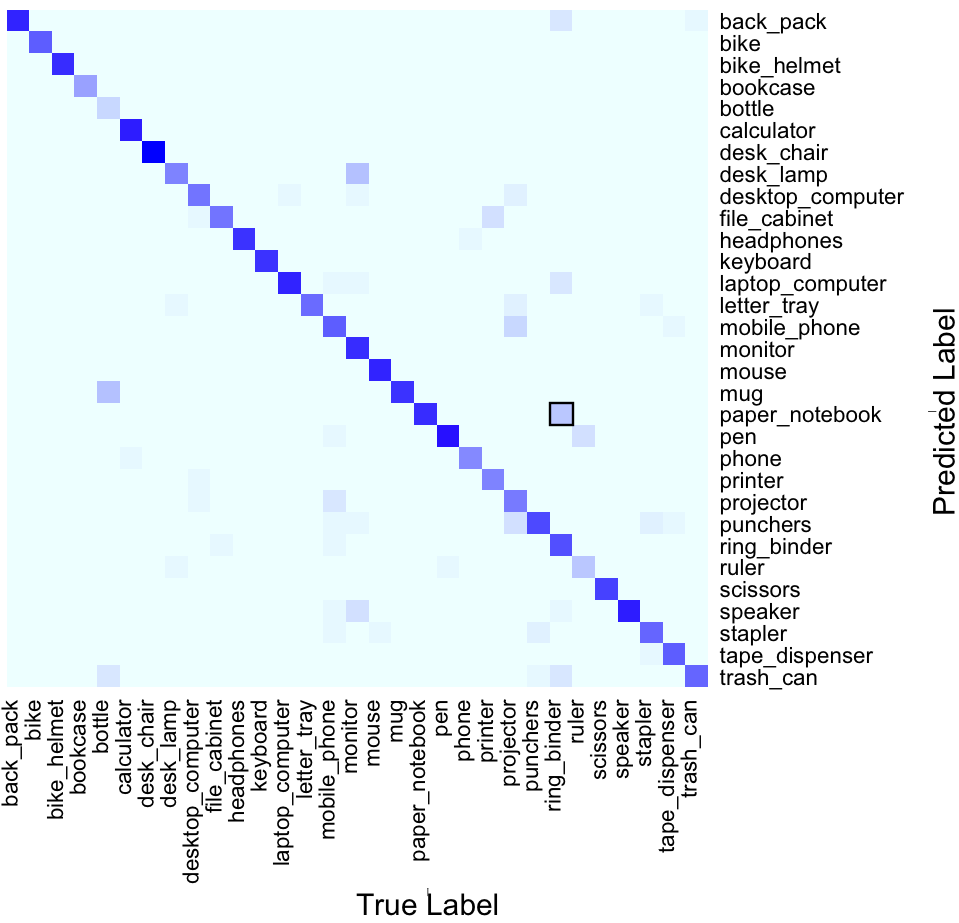}}
\caption{Confusion matrix of MADA}
\label{fig:3:2:3}
\end{subfigure}% 
\caption{Visualization for characterized class relationships and confusion matrix for task $\mathbf{A} \rightarrow \mathbf{W}$}
\label{fig:3:2}
\end{figure*}

\subsection{Main results}
The mean classification accuracy on \textit{ImageCLEF-DA} and \textit{Office-31} is reported in Table~\ref{tab:1} and Table~\ref{tab:2}. The results of baseline methods are reprinted from previous literatures \cite{pei2018multi,long2018conditional,jada2019,zhang2018collaborative,cao2018partial}.
RADA$_{d\rightarrow y}$ and RADA$_{y\rightarrow d}$ are our methods trained with $D_{KL}(\mathbf{\Omega_y} || \mathbf{\Omega_d})$ and $D_{KL}(\mathbf{\Omega_d} || \mathbf{\Omega_y})$.
As shown in Table~\ref{tab:1}, RADA$_{d\rightarrow y}$ and RADA$_{y\rightarrow d}$ both outperform baseline methods across all the transfer tasks for both \textit{ImageCLEF-DA} and \textit{Office-31}.  RADA$_{d\rightarrow y}$ slightly outperforms RADA$_{y\rightarrow d}$. This very similar performance suggests the choice of the target matrix does not make much difference in spite of the asymmetry of the adopted discrepancy metric, thus either of these two forms can be used in the RADA algorithm. The number of parameters used by RADA is also reduced to a large extent, compared with the multiple discriminators method (e.g. MADA). For \textit{ImageCLEF-DA} and \textit{Office31} datasets, the adoption of $12$ and $31$ independent two-layer discriminators $(x \rightarrow 1024 \rightarrow 1024 \rightarrow 1)$ generates more than $10^7$ parameters, whereas RADA only generates $\sim 10^6$ parameters. This reduction is important in practice, especially when the number of classes is very large (e.g. $>100$). The improved performance with even simpler network highlights the significance of incorporating class relationships into the adversarial training process. The alignment of class relationships between label predictor and domain discriminator introduces more structure information from the label space to the adversarial training process, and efficiently promotes the learning of transferable representations for feature extractor. We also include an ablation study of our method on different modules in the appendix.

We additionally provide evaluations for \emph{partial domain adaptation} (PDA) problem, where the target label space is a subset of source label space. It is a new technical bottleneck, which is more challenging and practical than the standard domain adaptation, considering the outlier classes in the source domain can cause negative transfer when discriminating the target classes \cite{cao2018partial,pei2018multi}. To show the robustness of our method against PDA, we implement the evaluation in a benchmark experimental setup. From \textit{Office-31}, we use all the categories for the source domain and choose the ten categories shared with \textit{Caltech256} \cite{griffin2007caltech} for the target domain. Among all the transfer tasks, the source domain contains $31$ classes and the target domain has $10$ classes. From Table \ref{tab:3}, we can observe that RADA outperforms \emph{ResNet} and other general domain adaptation methods, especially on the tasks $\mathbf{A} \rightarrow \mathbf{W}$, $\mathbf{A} \rightarrow \mathbf{D}$, $\mathbf{W} \rightarrow \mathbf{A}$ and $\mathbf{D} \rightarrow \mathbf{A}$, which suggests it successfully avoids the negative transfer trap.

\subsection{Empirical analysis}
\paragraph{Feature visualization} In order to visualize the embedded data, we use t-SNE to project the feature representations after \textit{pool5} in \textit{ResNet-50} that are respectively trained with DANN, MADA and RADA to lower dimensional space. The two-dimensional map of embedded data in target domain from the transfer task $\mathbf{A} \rightarrow \mathbf{W}$ is visualized in Figure \ref{fig:3:1}, where the class information is also given by assigning data points with different colors and numerical labels in the plot. We observe that the embedded features from different classes are better separated in RADA and MADA when compared with DANN. Although MADA is able to separate most of the data points according to their class labels, several classes around the center are still mixed up, while RADA can better separate those points. By integrating the information of class relationships, RADA can better extract the features uniquely belonging to each class and capture the modes of the data distribution.

\begin{figure}
    \centerline{\includegraphics[width=0.5\linewidth]{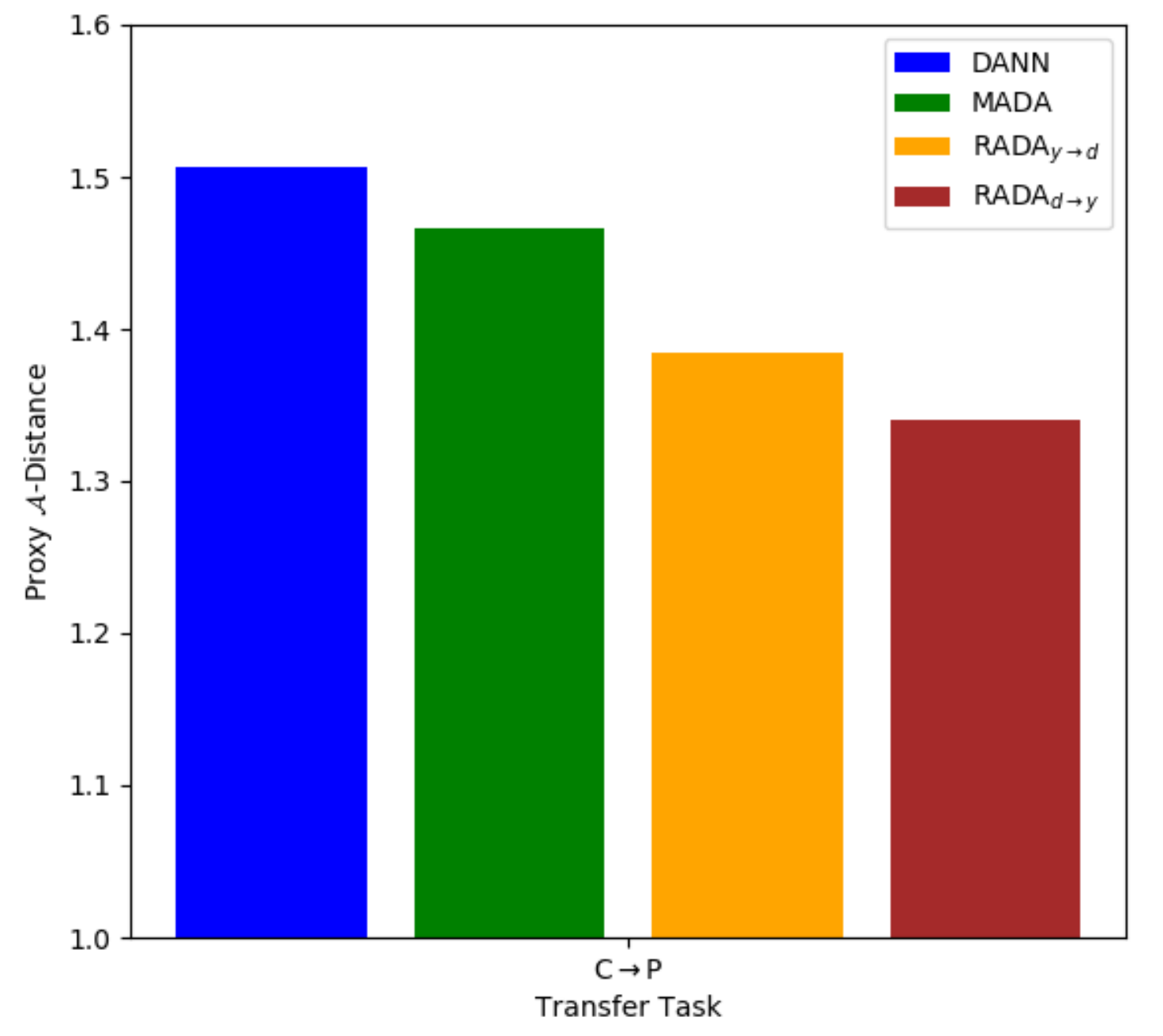}}
   \caption{Proxy $\mathcal{A}$-Distance}
   \label{fig:3:3}
\end{figure} 

\paragraph{Class relationships}
Partial correlation is a symmetric measure of association between two variables while controlling the effect of other variables. It is commonly used to model the conditional dependencies among a group of variables. The partial correlation between class $i$ and $j$ can be calculated by $\rho_{ij} = -\frac{\omega_{ij}}{\sqrt{\omega_{ii}\omega{jj}}}$ from the element $\omega_{ij}$ of precision matrix $\Omega$. With the estimated $\mathbf{\hat \Omega_y}$ and $\mathbf{\hat \Omega_d}$ from the transfer task $\mathbf{A} \rightarrow \mathbf{W}$ using RADA$_{d\rightarrow y}$, we calculate and visualize the partial correlations among all the classes in Figure \ref{fig:3:2}, where $\hat \rho_{ij}$ estimated from label predictor is displayed on the upper triangular part and that from the discriminator is on the lower triangular part. The symmetry of the heatmap in Figure \ref{fig:3:2:1} indicates our regularization successfully encourages the class relationships to be consistent between $G_d^*$ and $G_y$. Some class relationships are interesting and intuitive. For example, not surprisingly, the class \textit{paper notebook} is found to be positive associated with \textit{ring binder} from both label predictor and discriminator with RADA (black framed cell in Figure \ref{fig:3:2:1}). Aware of such class relationships, our method avoids miss-classifying several images (Figure \ref{fig:3:2:2}) of \textit{ring binder} as \textit{paper notebook} compared to MADA (black framed cell in Figure \ref{fig:3:2:3}).

\paragraph{Distribution discrepancy} Proxy $\mathcal{A}$-Distance (PAD) \cite{ben2007analysis,ganin2016domain} is a widely used metric to measure the feature distributional discrepancy between source and target domains.
PAD is defined as $d_{\mathcal{A}}=2(1-2\epsilon)$, where $\epsilon$ is the classification error (e.g. mean absolute error) of a domain classifier (e.g. SVM).
Generally, a lower PAD indicates a better generalization ability. As shown in Figure~\ref{fig:3:3}, on the transfer task $\mathbf{C}\rightarrow \mathbf{P}$, RADA outperforms DANN and MADA. This indicates RADA can better extract domain-invariant features. In addition, $d_{\mathcal{A}}$ of $\text{RADA}_{d\rightarrow y}$ are slightly lower than $\text{RADA}_{y\rightarrow d}$, showing that $\text{RADA}_{d\rightarrow y}$ has a better generalization ability.

\section{CONCLUSION}
  We present a novel approach to domain adaptation through revealing the structure information from the label space for aligning complicated data distributions during adversarial training. We propose a new design of multi-class domain discriminator and a novel regularizer to align the inter-class dependencies respectively characterized from label predictor and domain discriminator. Experiments show considering class relationship information can substantially improve the transfer learning performance. In this work, we model the class relationships in terms of symmetric relationships, and we will consider extending to a model that allows the class relationships to be asymmetric in our future study.

\section{APPENDIX}
\subsection{Detailed derivation of solution to equation (5)}
With the cyclic property of the trace, we have an exponential form of the objective function as:
\begin{equation}
  \begin{aligned}
  & \min_{\mathbf{\Omega_y}}- d_y\logdet(\mathbf{\Omega_y}) + \Tr(\mathbf{W_y}^{[L]}\mathbf{\Omega} {\mathbf{W_y}^{[L]}}^T) \\
  \Leftrightarrow &  \min_{\mathbf{\Omega_y}}- \det(\mathbf{\Omega_y})^{d_y}\exp(-\Tr(\mathbf{\Omega_y} {\mathbf{W_y}^{[L]}}^T\mathbf{W_y}^{[L]})) \\
    \end{aligned}
  \label{eqs:1}
  \end{equation}
 Let us denote $\mathbf{H} = \mathbf{{W_y}^{[L]}}^T\mathbf{W_y}^{[L]}$, which is assumed as a positive definite matrix, we have:\\
  \begin{equation}
  \begin{aligned}
  & \min_{\mathbf{\Omega_y}}- \det(\mathbf{\Omega_y})^{d_y}\exp(-\Tr(\mathbf{\Omega_y} \mathbf{H})) \\ 
  \end{aligned}
  \label{eqs:2}
  \end{equation}
  According to the spectral theorem in linear algebra, when $\mathbf{H}$ is a positive-definite symmetric matrix, then it  has precisely one positive-definite symmetric square root which we denote as $\mathbf{H}^{\frac{1}{2}}$. Thus with the cyclic property of the trace, equation (\ref{eqs:2}) can be rewritten to:
    \begin{equation}
  \begin{aligned}
  & \min_{\mathbf{\Omega_y}}- \det(\mathbf{\Omega_y})^{d_y}\exp(-\Tr(\mathbf{H}^{\frac{1}{2}}\mathbf{\Omega_y}\mathbf{H}^{\frac{1}{2}})) \\ 
  \end{aligned}
  \label{eqs:3}
  \end{equation}
  
Let $\mathbf{U} = \mathbf{H}^{\frac{1}{2}}\mathbf{\Omega_y}\mathbf{H}^{\frac{1}{2}}$ and solve:
\begin{equation}
  \begin{aligned}
&  \min_{\mathbf{U}}- \det(\mathbf{H})^{-d_y}\det(\mathbf{U})^{d_y}\exp(-\Tr(\mathbf{U})) \\ 
\Leftrightarrow &   \min_{\mathbf{U}}- \det(\mathbf{U})^{d_y}\exp(-\Tr(\mathbf{U})) \\
  \end{aligned}
  \label{eqs:4}
  \end{equation}
Let $\eta_1, \cdots, \eta_K \geq 0$ be the eigenvalues of matrix $\mathbf{U}$, then we have $\det(\mathbf{U}) = \prod_i^K{\eta_i}$ and $\Tr(\mathbf{U}) = \sum_i^K{\eta_i}$. Equation (\ref{eqs:4}) reduces to the problem of finding $\eta_i$ that:
\begin{equation}
  \begin{aligned}
\min_{\eta_i}- {\eta_i}^{d_y}\exp(-\eta_i), \forall i \\ 
  \end{aligned}
  \label{eqs:5}
  \end{equation}
With calculus, it is easy to get $\forall i$,  $\eta_i = d_y$. Assuming $\mathbf{V}$ is the matrix of eigen vectors of $\mathbf{U}$, we have:\\
$\mathbf{U}=\mathbf{V}(d_y\mathbf{I}_{K})\mathbf{V}^{{-1}}=d_y\mathbf{I}_{K}$
Finally we get:
\begin{equation}
  \begin{aligned}
 & \mathbf{H}^{\frac{1}{2}}\mathbf{\Omega_y}\mathbf{H}^{\frac{1}{2}} = d_y\mathbf{I}_{K}\\
\Rightarrow & \mathbf{\Omega_y} = ({\mathbf{H}^{-1}})^{\frac{1}{2}}d_y\mathbf{I}_{K}({\mathbf{H}^{-1}})^{\frac{1}{2}} \\
\Rightarrow & \mathbf{\Omega_y} = d_y\mathbf{H}^{-1} = d_y{({\mathbf{W_y}^{[L]}}^T \mathbf{W_y}^{[L]}})^{-1}\\
  \end{aligned}
  \label{eqs:52}
  \end{equation}

\subsection{Ablation study}
In order to provide an ablation study of our models, we conducted experiments with the same experimental setting on \textit{Office-31} that dropped the regularization item $L_R$ from our method to examine the effect of aligning inter-class dependencies in addition to using the single multi-class domain discriminator. The ablation study is reported in Table \ref{tab:s}, where the use of the single multi-class domain discriminator without the regularization item $L_R$ is noted as \emph{Only Multi-class Discriminator}. We find that the only use of a multi-class discriminator is better than MADA, and both RADA$_{d\rightarrow y}$ and RADA$_{y\rightarrow d}$ outperform the utilization of only a multi-class discriminator, implying that the alignment of class relationships can effectively improve the adaptation performance. \\
\begin{table}
\begin{center}
{\caption{Ablation study: mean accuracy (\%) on  \textit{Office-31} for UDA (\textit{ResNet-50})}
\label{tab:s}}
\begin{tabular}{ccccc}
Method & MADA & \makecell{\emph{Only Multi-class}\\ Discriminator} & RADA$_{y\rightarrow d}$ & RADA$_{d\rightarrow y}$\\
\hline
A$\rightarrow$W & 90.0 & 90.0 & 91.5 & 91.5\\
D$\rightarrow$W & 97.4 & 99.0 & 99.0 & 98.9\\
W$\rightarrow$D & 99.6 & 100.0 & 100.0 & 100.0\\
A$\rightarrow$D & 87.8 & 88.9 & 90.3 & 90.7 \\
D$\rightarrow$A & 70.3 & 70.5 & 71.5 & 71.5 \\
W$\rightarrow$A & 66.4 & 68.5 & 70.1 & 71.3 \\
Average & 85.2 & 86.2 & 87.1 & 87.3 \\
\hline
\end{tabular}
\end{center}
\end{table}

\bibliography{reference}
\end{document}